\documentclass[10pt,twocolumn,a4paper]{article}
\usepackage[margin=0.7in,columnsep=18pt]{geometry}
\usepackage[T1]{fontenc}
\usepackage[utf8]{inputenc}
\usepackage{lmodern}
\usepackage{microtype}
\usepackage{graphicx}
\usepackage{booktabs}
\usepackage{array}
\usepackage{tabularx}
\usepackage{multirow}
\usepackage{caption}
\usepackage{enumitem}
\usepackage{xcolor}
\usepackage{pifont}
\usepackage{amsmath}
\usepackage{float}
\usepackage{stfloats}
\usepackage[numbers,sort&compress]{natbib}
\usepackage[hidelinks]{hyperref}

\setlist[itemize]{nosep,leftmargin=1.4em}
\setlist[enumerate]{nosep,leftmargin=1.6em}
\newcolumntype{L}{>{\raggedright\arraybackslash}X}
\newcolumntype{R}{>{\raggedleft\arraybackslash}p{1.6cm}}
\newcommand{\tabsize}{\footnotesize}
\definecolor{bestgreen}{HTML}{1B7A3D}
\definecolor{failred}{HTML}{A6472B}
\newcommand{\best}[1]{\textbf{\textcolor{bestgreen}{#1}}}
\newcommand{\tnote}[1]{\par\smallskip\raggedright\footnotesize #1}
\newcommand{\yes}{\textcolor{bestgreen}{\ding{51}}}
\newcommand{\no}{\textcolor{failred}{\ding{55}}}

\graphicspath{{figures/}}

\title{\textbf{Configurable Multi-Stage Vision Pipeline for\\Crop Disease and Pest Diagnosis}}
\author{%
Naga Ganesh\textsuperscript{*} \and
Chandrashekar M S\textsuperscript{*} \and
Lakshmi Pedapudi\textsuperscript{\dag} \and
Aakash Singh \and
Vineet Singh\\[4pt]
\normalsize Digital Green}
\date{}

\begin{document}

\makeatletter
\twocolumn[%
  \begin{@twocolumnfalse}
  \maketitle
  \begingroup\footnotesize
  \begin{center}
  \textsuperscript{*}Equal contribution.\quad\textsuperscript{\dag}Corresponding author.\par
  \smallskip
  naga@digitalgreen.org, chandrashekar@digitalgreen.org, laxmigenius@gmail.com,
  aakash@digitalgreen.org, vineet.vinsing@gmail.com
  \end{center}
  \endgroup
  \begin{abstract}
  FarmerChat is Digital Green's farm advisory service for smallholder farmers. When something looks
wrong with a crop, farmers usually send a photograph as their entire query: no symptoms described,
no crop named, and often no text at all. The system must determine whether the image is usable,
identify the crop, and diagnose the disease or pest from photographs taken on cheap phones in real
field conditions. The current production system offers little control over these decisions:
quality thresholds cannot be adjusted, new crops and problems cannot be added, and there is no
configurable confidence threshold or fallback when the diagnosis is uncertain.

We study about 1.16 million photographs sent to FarmerChat from Ethiopia, India, Kenya, and
Nigeria. The production quality gate rejected 46.8\% of the images it judged, more than a quarter
of images reaching diagnosis received no crop label, and 35.8\% of labelled problems classified as
``disease'' were pests. We therefore split diagnosis into three independently evaluated stages:
image quality (M0), crop detection (M1), and disease or pest detection (M2). We evaluate two
routes: Route~A uses a single fine-tuned vision-language model (Qwen3-VL-4B), while Route~B uses
smaller specialist models (DaViT and YOLO26). Each stage can be replaced independently and its
thresholds can be configured.

We replace the production GPT-4o quality gate with a MobileNetV3 gate that reaches 86.9\% F1 with
12~ms latency. On a common test set, hierarchical DaViT-Base identifies crops with 95.41\%
accuracy, compared with 91.46\% for the production baseline. The same backbone also leads on
disease and pest identification and does not decline to answer, while the language models leave a
substantial share of rows without a diagnosis. The specialist route also has lower hosting cost at
the measured query volume.

The fine-tuned VLM provides two capabilities that the specialist route does not: it handles all
three stages in a single call and can request a more informative photograph when the available
image is insufficient for diagnosis.

  \end{abstract}
  \bigskip
  \end{@twocolumnfalse}
]
\makeatother

\section{Introduction}
\label{sec:intro}

Farmers using FarmerChat \citep{singh2024farmerchat} mostly report crop health problems by sending a photograph. These photographs look nothing like the tidy images in research datasets. The light is poor, the camera moves, and the subject changes from one photo to the next: a single leaf fills one frame, and the next holds a whole field, a hand, or a farm animal. Giving a useful answer therefore means making five decisions, not one:
\begin{enumerate}
  \item Is the image usable?
  \item What crop is it?
  \item What disease or pest, if any, is present?
  \item How confident is that call?
  \item What structured output does the downstream advisory system need?
\end{enumerate}

The system in production lets us change almost nothing. It offers none of the following:
\begin{itemize}
  \item adjustable thresholds for photograph rejection;
  \item addition of new crops, diseases or pests;
  \item control over how the model behaves, or a confidence cut-off;
  \item a choice of what happens when the answer is weak: reject it, try again, or send it to a person.
\end{itemize}
A system we cannot adjust throws away photographs a tunable one would keep, and it cannot be pointed at the crops and problems that matter in one country. Section~\ref{sec:failure} measures each of these on real farmer queries.

Table~\ref{tab:rq} lists the five questions this paper addresses.

\begin{table}[tb]
\centering\tabsize
\caption{Research questions.}
\label{tab:rq}
\begin{tabularx}{\linewidth}{@{}l L@{}}
\toprule
ID & Question \\
\midrule
RQ1 & Can a configurable local quality stage match the paid gate's decisions at lower cost and latency? \\
RQ2 & Does separating crop detection from disease and pest detection improve diagnosis? \\
RQ3 & How does a fine-tuned VLM compare with orchestrated specialist models? \\
RQ4 & What do per-country submission patterns imply for crop and problem coverage? \\
RQ5 & What is the accuracy, coverage, cost and latency trade-off? \\
\bottomrule
\end{tabularx}
\end{table}

This paper contributes four things:
\begin{itemize}
  \item A failure analysis of about 1.16 million farmer photographs from four countries.
  \item A MobileNetV3 quality gate that matches the paid gate's decisions at low memory and latency.
  \item A case for splitting pest detection from disease detection, and a four-head model that does it, released as a trained checkpoint.\textsuperscript{1}
  \item A benchmark of seven systems, specialist computer-vision models and vision-language models, on one test set. The four-head benchmark's labels and label space are released with it.\textsuperscript{2}
\end{itemize}

\begin{figure*}[b]
\footnotesize
\noindent\rule{0.3\columnwidth}{0.4pt}\par\vspace{3pt}
\noindent\textsuperscript{1}Checkpoint: \url{https://huggingface.co/DigiGreen/crop-disease-pest-detection-dg}\par
\noindent\textsuperscript{2}Benchmark labels: \url{https://huggingface.co/datasets/DigiGreen/Crop-Disease-Image-Eval-Synthetic}
\end{figure*}

\section{Background, Production Failures and Limitations}
\label{sec:existing}
\label{sec:failure}

\begin{figure*}[t]
\centering
\includegraphics[width=\linewidth]{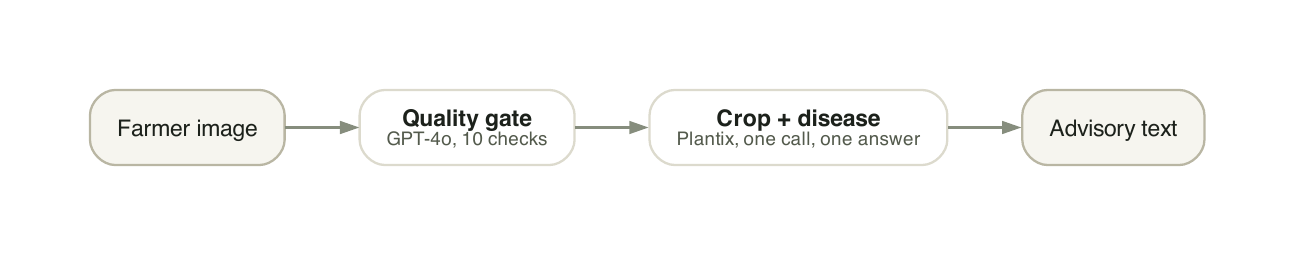}
\caption{Existing production path: two paid calls, no intermediate stage exposed to configuration.}
\label{fig:existing}
\end{figure*}

Figure~\ref{fig:existing} shows the production pipeline and Table~\ref{tab:existingfacts} summarises its stages. A farmer's photograph first goes through a GPT-4o quality gate, which performs ten checks in a single call. Images that pass are then sent to Plantix, which returns the crop and disease or pest in one response. Neither stage exposes settings that the caller can change.

\begin{table*}[t]
\centering\tabsize
\caption{Existing production path: stage facts from five production traces recorded on one day in August 2026.}
\label{tab:existingfacts}
\begin{tabularx}{\linewidth}{@{}l L@{}}
\toprule
Property & Value \\
\midrule
Quality gate & GPT-4o, one call per image, ten checks; \$0.0106 to \$0.0132 per image across the five traces \\
Diagnosis & Plantix, one call per image; crop and disease returned in one answer; 81-crop vocabulary \\
Caller controls & None. The gate returns pass or fail per check and Plantix returns a likelihood word (unlikely, likely, very likely); no threshold or numeric score can be set or read \\
Server-side latency & Quality gate 6.2 to 8.2~s, Plantix diagnosis 2.8 to 4.4~s, advisory text 1.5 to 2.9~s, total 10.5 to 13.7~s per image across the five traces; the quality gate is the slowest stage in every trace \\
\bottomrule
\end{tabularx}
\end{table*}

\begin{itemize}
  \item Neither stage has a documented or adjustable threshold.
  \item The paid quality-gate call is the slowest stage in every measured trace (Table~\ref{tab:existingfacts}).
\end{itemize}

Plant disease recognition from images has been studied extensively \citep{mohanty2016using}, using datasets such as PlantVillage \citep{hughes2015plantvillage} and field datasets such as PlantDoc \citep{singh2020plantdoc}. These datasets also show the gap between controlled images and photographs taken in real field conditions. Our setting differs in three ways. First, our images are photographs submitted by farmers to a live service and were not curated for this study. Second, our labels come from a panel of models rather than experts labelling images at scale. Third, we study how the diagnosis pipeline can be configured and improved, rather than the accuracy of a single classifier.

We use established architectures for the proposed pipeline: MobileNetV3 \citep{howard2019mobilenetv3} for the quality gate, DaViT \citep{ding2022davit} and YOLO \citep{ultralytics_yolo26} for the specialist route, and Qwen3-VL \citep{qwen3vl} and Gemma~3 \citep{gemma3} as fine-tuning candidates. The FarmerChat platform is described in \citet{singh2024farmerchat}. Our evaluation approach also builds on our earlier work on conversational-AI evaluation \citep{singh2026conversational} and agricultural ASR benchmarking \citep{pedapudi2026asr}.

Unless otherwise stated, the measurements in this paper cover all photographs sent to FarmerChat up to August 2026. Each table specifies the rows included in its analysis.

\subsection{Production Failures}
The three subsections below measure what the existing pipeline does with the photographs it receives: which images it throws away, which crops it can name, and which problems it can name.

\subsubsection{Image Quality Failures}
\label{sec:fail-quality}
\begin{figure*}[t]
\centering
\setlength{\tabcolsep}{3pt}
\begin{tabular}{@{}ccc@{}}
\includegraphics[width=0.315\linewidth]{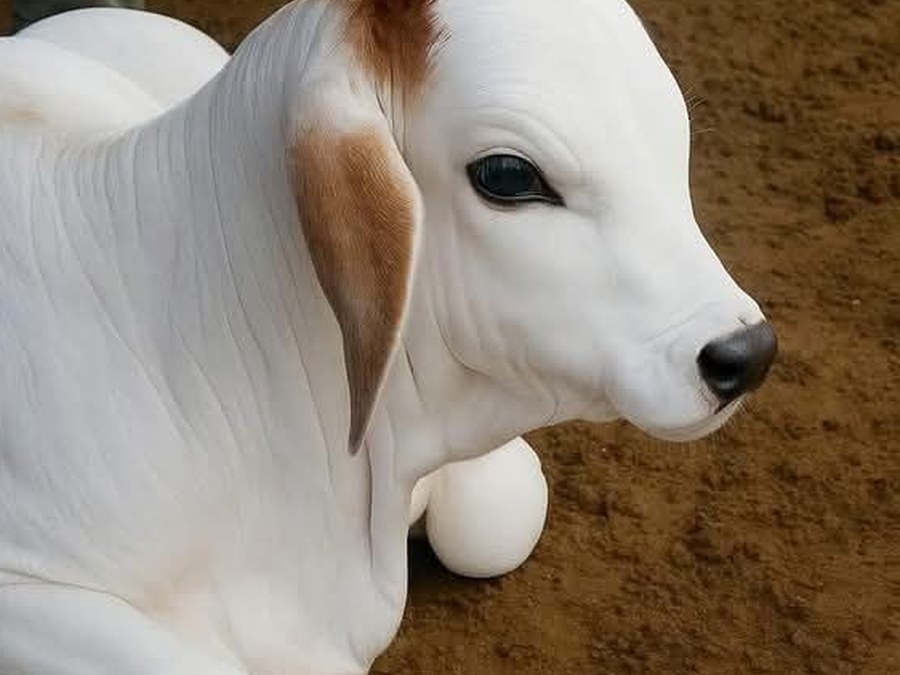} &
\includegraphics[width=0.315\linewidth]{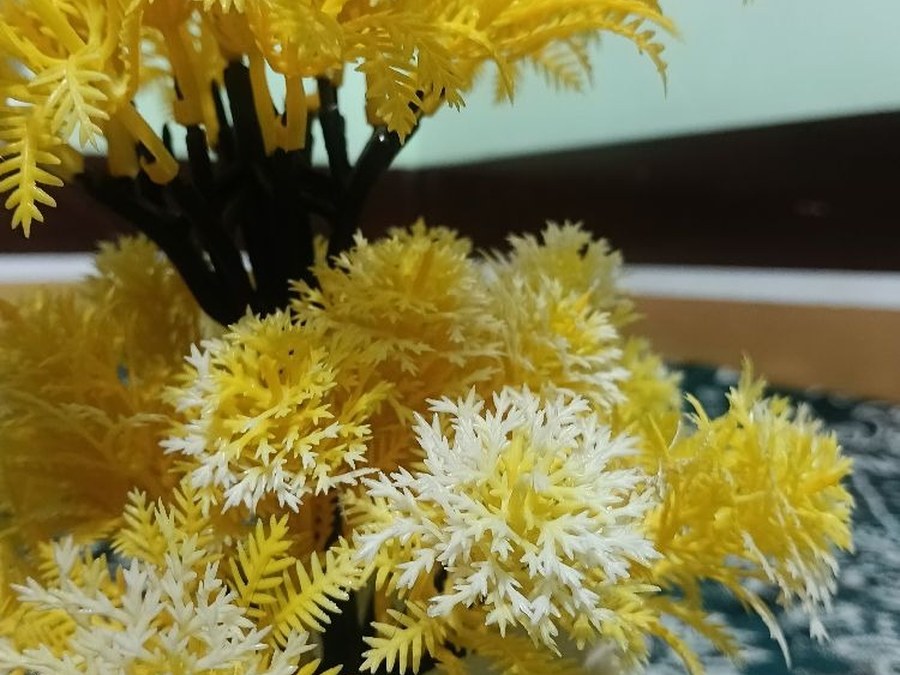} &
\includegraphics[width=0.315\linewidth]{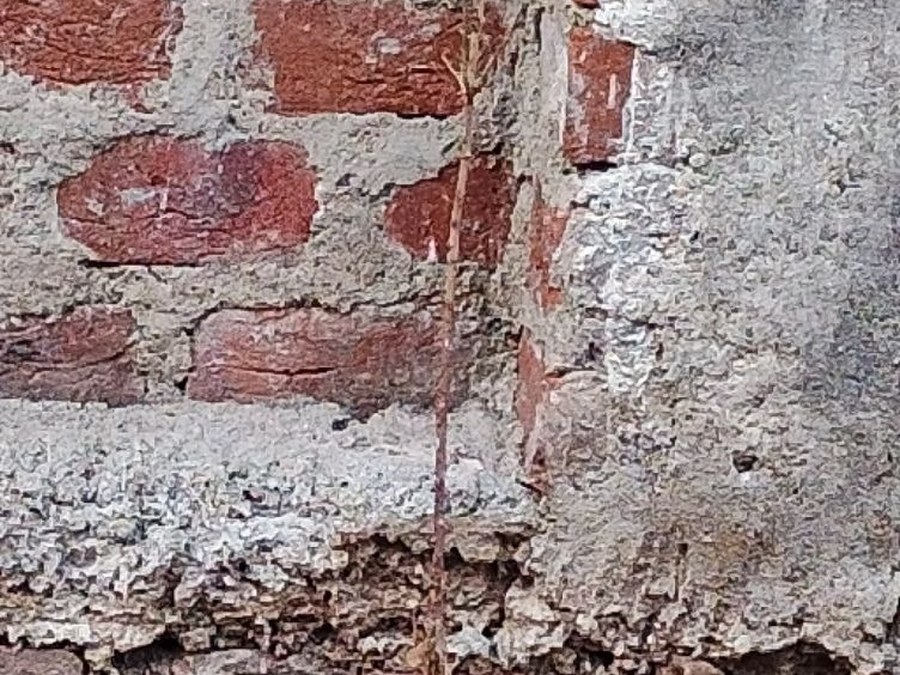} \\
\footnotesize\textbf{Plant Detected} & \footnotesize\textbf{Plant Detected} & \footnotesize\textbf{Dominant Content} \\
\footnotesize A model animal, not a plant & \footnotesize Artificial flowers indoors & \footnotesize A wall fills the frame \\[5pt]
\includegraphics[width=0.315\linewidth]{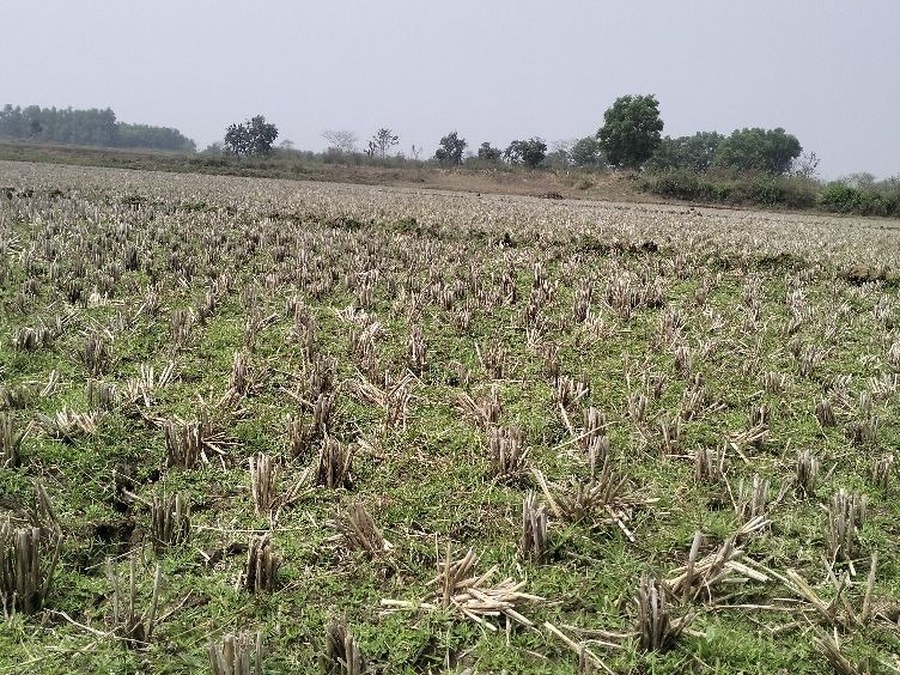} &
\includegraphics[width=0.315\linewidth]{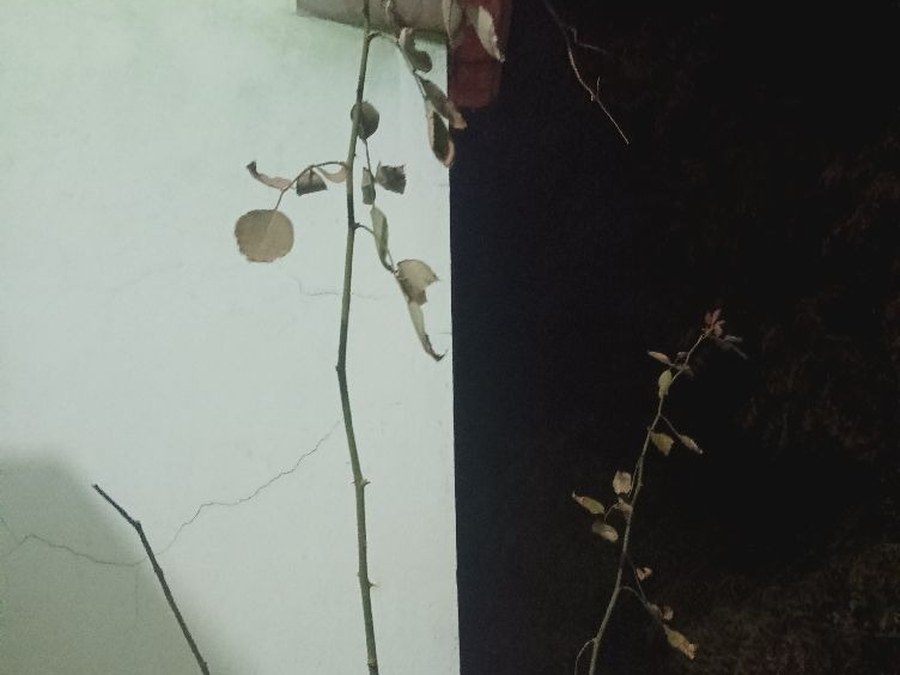} &
\includegraphics[width=0.315\linewidth]{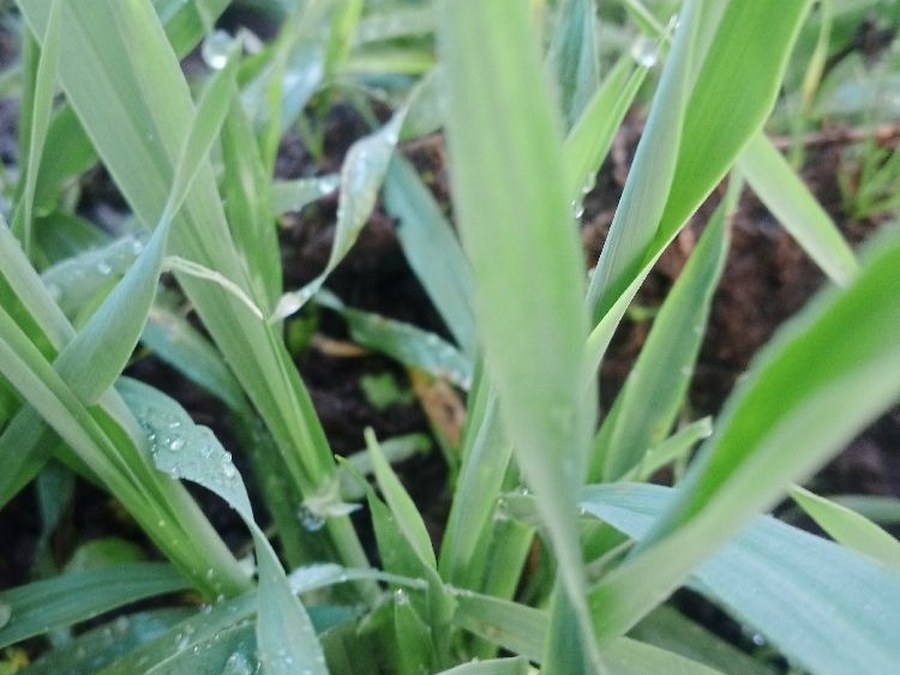} \\
\footnotesize\textbf{Dominant Content} & \footnotesize\textbf{Lighting} & \footnotesize\textbf{Focus} \\
\footnotesize A whole field, no subject & \footnotesize Half the frame in shadow & \footnotesize Too soft to read a leaf \\
\end{tabular}
\caption{Photographs the quality gate rejected, across four of the ten checks.}
\label{fig:rejects}
\tnote{Held-out quality-gate test split of Table~\ref{tab:splits}. Each image failed exactly one of the ten checks, so the label names that check.}
\end{figure*}

Table~\ref{tab:gatefunnel} follows the photographs through the pipeline. The gate judged seven in ten of them and rejected nearly half of those it judged. Most rejected photographs were never sent to Plantix, though some were sent anyway. Table~\ref{tab:qfail} gives the fail rate of each check. Figure~\ref{fig:rejects} shows what a rejected photograph actually looks like: six photographs across four of the ten checks.

\begin{table*}[t]
\centering\tabsize
\caption{Submission funnel over all photographs submitted to FarmerChat up to August 2026. Each share names its denominator.}
\label{tab:gatefunnel}
\begin{tabularx}{\linewidth}{@{}L r l@{}}
\toprule
Stage & Images & Share \\
\midrule
Submitted & 1,163,658 & \\
No quality decision recorded & 341,871 & 29.4\% of submitted \\
Judged by the gate & 821,787 & 70.6\% of submitted \\
\quad Rejected & 384,271 & 46.8\% of judged \\
\quad\quad Withheld from the diagnosis call & 328,520 & 85.5\% of rejected; 28.2\% of submitted \\
\quad\quad Sent for diagnosis anyway & 55,751 & 14.5\% of rejected \\
Sent for diagnosis & 833,681 & 71.6\% of submitted \\
\quad Crop named & 608,742 & 73.0\% of sent \\
\quad No crop named & 224,939 & 27.0\% of sent \\
\bottomrule
\end{tabularx}
\end{table*}

\begin{table}[tb]
\centering\tabsize
\caption{How often each quality check fails.}
\label{tab:qfail}
\begin{tabular}{@{}llr@{}}
\toprule
Check & Type & Fail rate \\
\midrule
Dominant Content & semantic & 35.7\% \\
Resolution & photometric & 32.8\% \\
Focus & photometric & 32.2\% \\
Obstruction & semantic & 27.7\% \\
Motion Blur & photometric & 27.6\% \\
Plant Detected & semantic & 26.0\% \\
Lighting & photometric & 25.3\% \\
Orientation & semantic & 25.2\% \\
Color Balance & photometric & 21.6\% \\
Noise & photometric & 20.5\% \\
\bottomrule
\end{tabular}
\tnote{On the quality-gate training split of Table~\ref{tab:splits}, half gate-rejected images and half gate-accepted. The labels are the GPT-4o gate's own.}
\end{table}

\begin{itemize}
  \item The highest-volume check, Dominant Content, is semantic (``what is in frame''), as is Plant Detected. Blur and contrast statistics cannot assess either check, which is why \S\ref{sec:m0} uses a learned model rather than calibrated thresholds alone.
  \item Separating usable images that were wrongly rejected from the rejection rate as a whole needs its own measurement, listed in \S\ref{sec:limits}.
\end{itemize}

\subsubsection{Crop Coverage}
\label{sec:fail-crop}
\begin{figure*}[t]
\centering
\includegraphics[width=\linewidth]{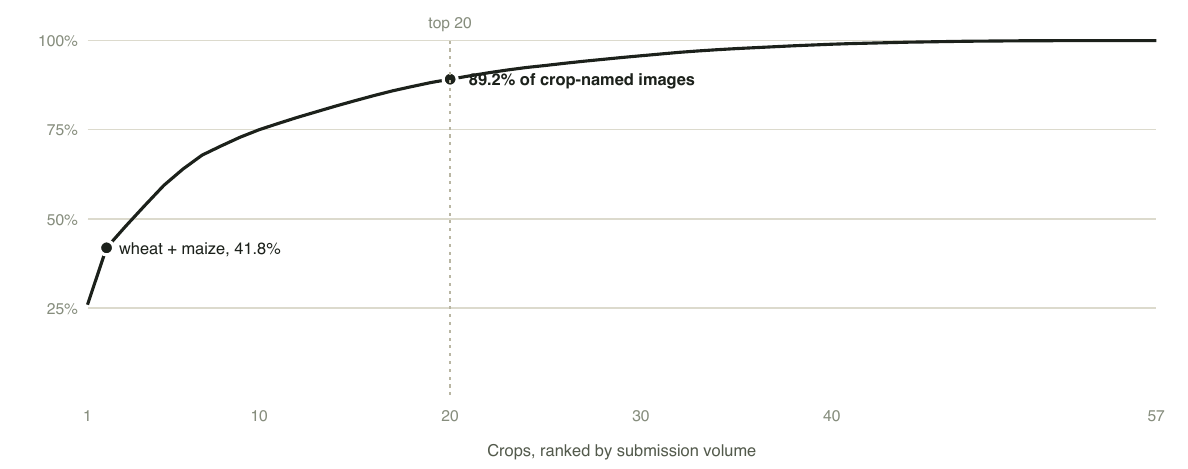}
\caption{Cumulative coverage of crop-named images by the top-N crops, ranked by volume.}
\label{fig:coverage}
\end{figure*}

Of the photographs sent for diagnosis, more than a quarter came back with no crop named (Table~\ref{tab:gatefunnel}). Among those that did get a crop, Table~\ref{tab:crops} lists the ten sent most often. Figure~\ref{fig:coverage} shows how much of the farmer query volume the top crops cover.

\begin{table}[tb]
\centering\tabsize
\caption{Top crops by volume with cumulative share of the 608,742 crop-named images.}
\label{tab:crops}
\begin{tabular}{@{}rlrr@{}}
\toprule
Rank & Crop & Images & Cumulative share \\
\midrule
1 & Wheat & 157,526 & 25.9\% \\
2 & Maize & 97,176 & 41.8\% \\
3 & Cabbage & 36,933 & 47.9\% \\
4 & Rice & 35,280 & 53.7\% \\
5 & Potato & 34,566 & 59.4\% \\
6 & Tomato & 27,728 & 63.9\% \\
7 & Pepper & 23,768 & 67.8\% \\
8 & Coffee & 15,945 & 70.5\% \\
9 & Bean & 14,839 & 72.9\% \\
10 & Banana & 12,715 & 75.0\% \\
\bottomrule
\end{tabular}
\end{table}

\subsubsection{Disease and Pest Coverage}
\label{sec:fail-disease}
We sorted every label in the problem vocabulary into a type using keyword rules (Table~\ref{tab:probtype}). More than a third of the labelled problems are pests, not pathogens. Rare classes are also poorly represented: 116 of the 250 disease classes in the evaluation label set have fewer than 10 test images. And the problems named most often mix insects and pathogens together without saying which is which (Table~\ref{tab:topcrops}).

\begin{table}[tb]
\centering\tabsize
\caption{Problem vocabulary by type.}
\label{tab:probtype}
\begin{tabular}{@{}lrr@{}}
\toprule
Problem type & Share of occurrences & Distinct labels \\
\midrule
Disease (pathogen) & 37.1\% & 1,548 \\
Pest (insect) & 35.8\% & 1,907 \\
Nutrient deficiency & 10.3\% & 153 \\
Virus & 9.1\% & 408 \\
Unclassified & 7.5\% & 1,632 \\
Weed & $<$0.1\% & 39 \\
\midrule
All types & 100\% & 5,689 \\
\bottomrule
\end{tabular}
\tnote{Shares are of the 401,594 labelled occurrences; the last column counts distinct canonical labels. Two labels marked unspecified are counted in the total only, and shares are rounded.}
\end{table}

\subsection{Limitations}
The measurements above point at six limitations of the system in production:
\begin{itemize}
  \item Quality thresholds are neither visible nor adjustable.
  \item Nearly half of the images the gate judged were rejected, and more than a quarter of all submissions never reached diagnosis (Table~\ref{tab:gatefunnel}).
  \item An 81-crop vocabulary: images showing crops outside this list receive no diagnosis.
  \item No per-stage failure visibility: a wrong answer cannot be traced to quality, crop or disease.
  \item No country-specific tuning.
  \item No structured path for a human correction to reach model training.
\end{itemize}

\subsubsection{Country-Level Funnel}
\label{sec:fail-country}
\begin{figure*}[t]
\centering
\includegraphics[width=\linewidth]{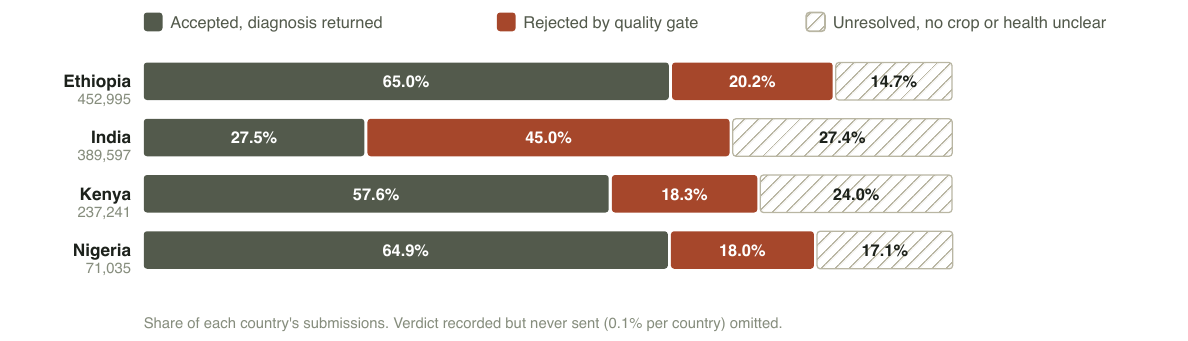}
\caption{Outcome shares by country.}
\label{fig:funnel}
\end{figure*}

Table~\ref{tab:funnel} and Figure~\ref{fig:funnel} follow the photographs through the pipeline for the four countries that send almost all of them. Table~\ref{tab:topcrops} shows what each country sends.

\begin{table*}[t]
\centering\tabsize
\caption{Submission outcomes for the four countries that send 98.9\% of all photographs.}
\label{tab:funnel}
\begin{tabularx}{\linewidth}{@{}l *{8}{>{\raggedleft\arraybackslash}X}@{}}
\toprule
Country & Total & Rejected & Rejected share & Crop named & Healthy & Healthy share & Disease & Livestock signal \\
\midrule
Ethiopia & 452,995 & 91,288 & 20.2\% & 300,117 & 206,201 & 68.7\% & 88,328 & 1,815 \\
India & 389,597 & 175,125 & 45.0\% & 116,521 & 38,057 & 32.7\% & 69,239 & 4,740 \\
Kenya & 237,241 & 43,374 & 18.3\% & 141,369 & 81,880 & 57.9\% & 54,796 & 4,192 \\
Nigeria & 71,035 & 12,772 & 18.0\% & 47,059 & 32,923 & 70.0\% & 13,163 & 344 \\
\bottomrule
\end{tabularx}
\tnote{Rejected share is of the country's own total, not of judged images as in Table~\ref{tab:gatefunnel}. Healthy share is of its crop-named images.}
\end{table*}

\begin{table*}[t]
\centering\tabsize
\caption{Top five crops and top three named problems by country.}
\label{tab:topcrops}
\begin{tabularx}{\linewidth}{@{}l L r L@{}}
\toprule
Country & Top five crops & Top-5 share & Top three named problems \\
\midrule
Ethiopia & wheat 51.4\%, maize 17.2\%, cabbage 7.1\%, potato 6.6\%, tomato 2.9\% & 85.2\% & Fall Armyworm 14.5\%, Fusarium Head Blight 13.7\%, Northern Leaf Blight 12.1\% \\
India & rice 11.6\%, pepper 10.6\%, cucumber 7.9\%, tomato 7.7\%, eggplant 7.2\% & 45.0\% & Nitrogen Deficiency 13.1\%, Tobacco Caterpillar 6.6\%, Potassium Deficiency 6.6\% \\
Kenya & maize 25.2\%, cabbage 12.0\%, potato 8.1\%, tomato 7.9\%, coffee 7.1\% & 60.3\% & Cercospora Leaf Spot of Beet 15.3\%, Fall Armyworm 13.7\%, Tobacco Caterpillar 6.2\% \\
Nigeria & rice 38.0\%, maize 14.4\%, millet 5.6\%, pepper 5.6\%, sorghum 5.2\% & 68.7\% & Nitrogen Deficiency 19.5\%, Fall Armyworm 10.9\%, Red Cotton Bug 8.4\% \\
\bottomrule
\end{tabularx}
\tnote{Crop shares are of that country's crop-named images; problem shares are of its problem-named images.}
\end{table*}

\begin{itemize}
  \item India's rejection rate is more than double Kenya's (Table~\ref{tab:funnel}). One flat quality threshold does not fit both.
  \item India's healthy share among crop-named images is under half of Ethiopia's (Table~\ref{tab:funnel}). What a ``reasonable'' disease rate looks like differs by country.
  \item India's five most-submitted crops cover under half of its crop-named images, against more than four-fifths in Ethiopia (Table~\ref{tab:topcrops}). A single crop vocabulary therefore fits some countries better than others.
  \item Two of the three most-named problems in India and Nigeria are missing nutrients, not diseases, and Fall Armyworm, an insect, is in the top three in three of the four countries. All of this is what one ``disease'' head is currently asked to cover.
\end{itemize}

\subsubsection{Human Review Labelling}
\label{sec:fail-review}
Human review is critical for building models that are reliable and grounded in real-world conditions. However, producing ground-truth labels from scratch is time-consuming and limits how much data can be reviewed. A more scalable approach is to have reviewers assess AI-annotated samples rather than generate every label independently. Instead of asking a reviewer to identify the crop and problem from a blank form, the model can propose an annotation that the reviewer verifies or corrects. This shifts human effort from generating labels to validating them, allowing substantially more samples to be reviewed within the same time and creating a larger pool of reliable labels for model evaluation and training. Section~\ref{sec:limits} describes how this human-in-the-loop approach can be incorporated into the pipeline.

\section{Pipeline Design}
\label{sec:arch}

Each principle in Table~\ref{tab:principles} answers a finding in \S\ref{sec:failure}.

\begin{table}[tb]
\centering\tabsize
\caption{Six design principles.}
\label{tab:principles}
\begin{tabularx}{\linewidth}{@{}l L@{}}
\toprule
ID & Principle \\
\midrule
P1 & Separate image quality from diagnosis. A diagnosis model is never asked to compensate for an unusable image. \\
P2 & Tune the quality gate to keep usable images: reject what cannot be read, and rarely reject what can. \\
P3 & Break the problem into separate decisions: quality, crop, disease, pest. \\
P4 & Make every threshold a setting of ours rather than a provider default we inherit. \\
P5 & Let production data decide coverage: the crops, countries, diseases, pests and photo conditions the pipeline actually sees. \\
P6 & Measure every stage independently, not only the final answer. \\
\bottomrule
\end{tabularx}
\end{table}

\subsection{Three-Stage Architecture}
\begin{figure*}[t]
\centering
\includegraphics[width=\linewidth]{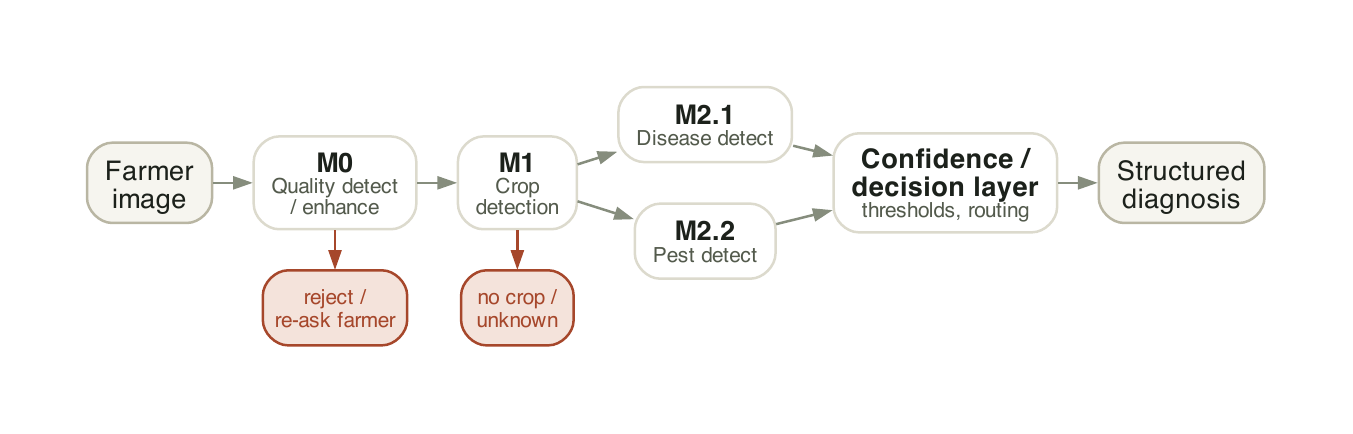}
\caption{Proposed three-stage pipeline with named reject branches.}
\label{fig:pipeline}
\end{figure*}

Figure~\ref{fig:pipeline} shows the proposed pipeline. M0 determines whether the photograph is usable and sends recoverable images for enhancement. M1 identifies the crop. M2.1 identifies the disease, while M2.2 identifies the pest. A final decision layer applies the configured thresholds and determines how each image is routed.

If an image is rejected by M0 or M1, the pipeline records the reason so the farmer can be asked for a better photograph instead of receiving no diagnosis. Both routes (\S\ref{sec:route-a}, \S\ref{sec:route-b}) use this same architecture; they differ only in whether the stages are implemented as separate model calls or as a single model call.

\subsection{Module M0: Image Quality Detection and Enhancement}
\label{sec:m0}

Let as many useful images through to M1 and M2 as possible, while staying fast and cheap.

\subsubsection{Quality Gate A: VLM Reference}
The production gate runs GPT-4o once per image with ten checks (Motion Blur, Lighting, Focus, Obstruction, Color Balance, Dominant Content, Resolution, Noise, Orientation, Plant Detected) at the per-image price in Table~\ref{tab:existingfacts}. We treat its answers as the labels Gate~B learns from, not as the truth: called twice on the same image it does not always say the same thing.

\subsubsection{Quality Gate B: Lightweight CV Model}
\label{sec:m0-gateb}
Gate~B is the replacement we propose: a small model trained to copy Gate~A's decisions, so that the threshold is ours to move and no paid call is made per image. We built five candidates, ranging from simple threshold rules to a convolutional network: calibrated threshold rules over blur, exposure and contrast statistics; gradient-boosted trees over the same statistics; a hybrid of those trees with a tiny convolutional network; a tiny custom CNN sized to fit on a phone; and MobileNetV3-small, a standard mobile backbone.

Each candidate was scored against Gate~A's answers on the quality-gate split of Table~\ref{tab:splits}, whose train, validation and test images never share an identifier, both overall and one check at a time. The metric is F1, which balances precision and recall for rejection: how often a rejection is correct and how many images that should be rejected are identified. The target, fixed before the runs, was 88\% F1. 

Two things are asked of the winner besides agreement with Gate~A: a median latency low enough to sit in front of every request, and a file small enough that the same gate could later run on the phone rather than on a server. Section~\ref{sec:res-m0} reports all three, per candidate and per check.

\subsubsection{Image Enhancement and Routing}
M0's answer sends an image one of three ways. A good image goes straight to M1. A fixable one is cleaned up first and then goes to M1. An unusable one is rejected, and the farmer is asked for a better photograph. The cleaning step is a proposal only: we have not tested any method for it yet (\S\ref{sec:limits}).

\subsection{Module M1: Crop Detection}
\label{sec:m1}

\begin{itemize}
  \item \textbf{Input:} a quality-passed image, plus the farmer's registered crop profile where one exists.
  \item \textbf{Output:} crop name, confidence, and an ``unknown or unsupported'' flag when confidence falls below the configured threshold.
\end{itemize}

Three models can fill M1. They differ on one thing: whether the crop is answered on its own or in the same call as the diagnosis.
\begin{itemize}
  \item \textbf{Qwen3-VL-4B, fine-tuned.} The Route~A model (\S\ref{sec:route-a}). Answers the crop in the same call as the diagnosis, in its own words rather than from a list. Scored in Table~\ref{tab:single}.
  \item \textbf{DaViT-Base, fine-tuned.} The Route~B reference model (\S\ref{sec:route-b}). Answers the crop from a fixed list, in the same pass as the category, disease and pest heads. Scored in Tables~\ref{tab:single} and \ref{tab:hier}.
  \item \textbf{YOLO26x-cls, fine-tuned.} Carries the four-part head of \S\ref{sec:m2}, so one model answers crop, category, disease and pest together. Scored on crop in Table~\ref{tab:single} and against DaViT-Base on all four heads in Tables~\ref{tab:hier} and \ref{tab:hiercost}.
\end{itemize}

Two of the three sort an image into a fixed list of classes and one answers in free text. None draws a box around the problem, and none can name a crop outside its list.

The ``unknown'' cut-off is a setting, not a learned value. Calibrating it against a target error rate is listed in \S\ref{sec:limits}.

\subsection{Module M2: Disease and Pest Detection}
\label{sec:m2}

\subsubsection{Two Sub-Modules}
\begin{itemize}
  \item \textbf{Disease detection (M2.1).} Input: image plus the crop label from M1. Output: disease name and confidence. Severity is out of scope here (\S\ref{sec:limits}).
  \item \textbf{Pest detection (M2.2).} Input: image plus the crop label from M1, taken as one more input rather than as a filter. Output: pest name, confidence.
\end{itemize}

\subsubsection{Conditional Routing}
\label{sec:m2-routing}
The system in production sends every image through the Plantix service after a quality gate. Plantix detects the crop and the disease or pest together, in one answer. Our design separates the two because crop-based narrowing helps disease detection but hurts pest detection. For pathogens narrowing matches the biology: the same disease looks different on different plants, so knowing the plant helps, and the disease list is narrowed to what occurs on that crop. For insects the plant helps less and narrowing hurts. A caterpillar looks like a caterpillar whatever plant it sits on, and a crop found on the wrong plant would rule the right insect out. So the crop label goes to both heads, and only the disease head is allowed to narrow its list by it. The pest head keeps all 92 choices open, which is what stops a wrong M1 call from losing the answer.

\begin{figure*}[t]
\centering
\includegraphics[width=0.92\linewidth]{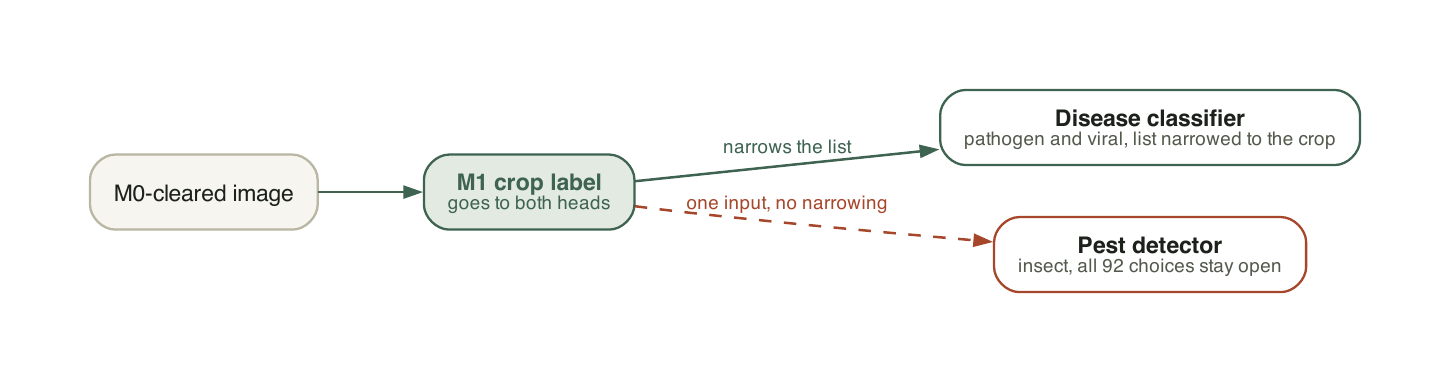}
\caption{Proposed M2 routing. The crop label reaches both heads and narrows only disease.}
\label{fig:m2}
\end{figure*}

 Figure~\ref{fig:m2} and Table~\ref{tab:m2routing} give the split we propose. How much of the query volume each branch carries is in Table~\ref{tab:probtype}: pathogens and viruses together are the larger share, insects the next.

\begin{table*}[t]
\centering\tabsize
\caption{Proposed M2 routing, replacing one Plantix call that returns crop and problem together. ``Needs'' means the route cannot run without the label.}
\label{tab:m2routing}
\begin{tabularx}{\linewidth}{@{}l l L@{}}
\toprule
Route & Needs M1 crop label & Model type \\
\midrule
Disease (pathogen and viral) & \yes & Per-crop classifier, unchanged \\
Pest (insect) & Optional & Unfiltered 92-way detector that may read the label, YOLO candidate \\
\bottomrule
\end{tabularx}
\tnote{The pest share is a row in Table~\ref{tab:probtype}; the disease branch is its pathogen and virus rows together.}
\end{table*}

\subsubsection{Reference Implementation}
The model in \S\ref{sec:res-hier} does exactly this split. One shared backbone feeds four heads:
\begin{itemize}
  \item \textbf{Crop}, 110 choices, and \textbf{category}, 3 choices. The category answer is what picks between the next two heads, and it is made in the same pass.
  \item \textbf{Disease}, 285 choices, filtered by a table of which diseases occur on which crop. It needs the crop first.
  \item \textbf{Pest}, 92 choices, not filtered at all.
\end{itemize}
This is one model with two separately gated heads, rather than two independent pipelines. In
this reference model the four heads share a backbone and the pest head is not handed the crop label
itself, so the optional input of Figure~\ref{fig:m2} is part of the proposed design and not of the
model scored in \S\ref{sec:res-hier}.

\subsection{Route A: Fine-Tuned Vision-Language Model}
\label{sec:route-a}

\begin{figure*}[t]
\centering
\includegraphics[width=\linewidth]{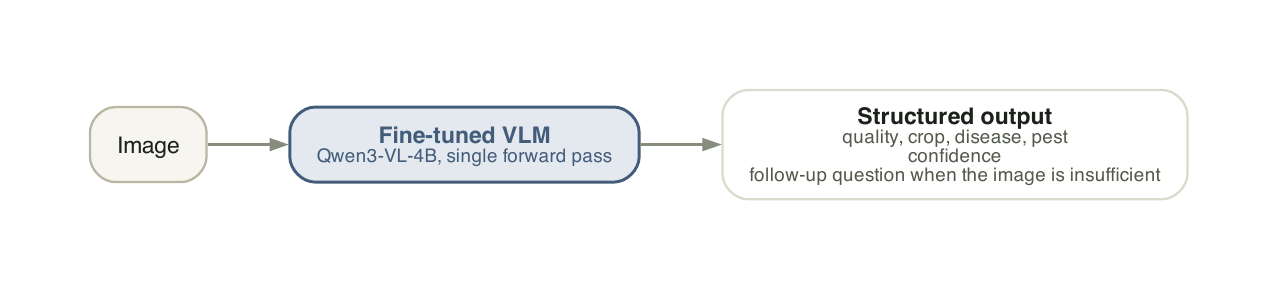}
\caption{Route A. One model, one call, no hand-off between stages.}
\label{fig:routea}
\end{figure*}

Figure~\ref{fig:routea} shows the route: one model answers all three stages in one call.

\begin{figure*}[t]
\centering
\begin{minipage}{0.93\linewidth}
\ttfamily\scriptsize
\begin{verbatim}
{ "analysis_status": "full",
  "crop": {
    "present": true, "primary_name": "maize", "scientific_name": "Zea mays",
    "growth_stage": "vegetative", "other_crops_visible": [] },
  "health_issue": {
    "present": true, "category": "pest", "specific_name": "fall armyworm",
    "severity": "moderate", "affected_area_percent_estimate": 30,
    "affected_parts": ["leaf", "whorl"],
    "symptoms_observed": [
      "ragged elongated feeding hole in leaf",
      "dark frass-like exudate with reddish-brown coloration",
      "abundant pale granular frass deposits around feeding site",
      "tissue chewed and torn at the whorl" ],
    "co_occurring_issues": [] },
  "image_quality": {
    "overall": "good", "issues": [],
    "diagnostic_usability": "yes", "authenticity": "real_photo" },
  "context": {
    "setting": "open_field", "camera_distance": "close",
    "soil_visible": true, "soil_condition": null },
  "recommended_followup_image":
    "close-up of the whorl interior to confirm presence of larvae
     and characteristic inverted-Y head capsule markings" }
\end{verbatim}
\end{minipage}
\caption{One Route A output, as returned. Every field is machine-readable, including the follow-up it asks for.}
\label{fig:record}
\end{figure*}

\subsubsection{Candidate Model}
Qwen3-VL-4B, fine-tuned on data curated from the panel-labelled images (\S\ref{sec:data}). Gemma-3-4B was considered alongside it; we chose Qwen3-VL-4B on the proof-of-concept results. None of the Qwen fine-tune's training images appears in the scoring set of \S\ref{sec:method-splits}, so its scores in \S\ref{sec:results} are out of sample. It also trained on almost the same images as Route~B, but none of the images used to validate or test Route~B were included, so the two routes are close to a matched comparison.

\subsubsection{What Fine-Tuning Changes}
A general-purpose model like Qwen3-VL-4B will attempt an answer even when the photograph does not provide enough information. The fine-tuned model, in contrast, can recognize when more information is needed and ask for a specific follow-up photograph. Figure~\ref{fig:record} shows one example, and Table~\ref{tab:followups} gives five more.
\begin{itemize}
  \item It asks for one of five things: a better photograph, a closer look at the crop, a closer look at the damaged part, whether the subject is an animal, or plain advice rather than another question.
  \item The request is specific to the crop and suspected problem: it tells the farmer what to photograph and what the new image should help confirm. This is more useful than simply asking for a clearer picture. For example, the model may ask for a close-up that helps distinguish between two insects that look similar but require different treatments (Table~\ref{tab:followups}).
  \item It also asks when it is already right: on many rows it named both the crop and the problem correctly and still asked for the photograph an agronomist would want before recommending treatment (\S\ref{sec:res-error}).
  \item Table~\ref{tab:tradeoff} carries how often it asks, measured on the shared test rows of \S\ref{sec:method-splits}.
\end{itemize}

\begin{table*}[t]
\centering\tabsize
\caption{Follow-up photographs the fine-tune asks for, taken from its own output.}
\label{tab:followups}
\begin{tabularx}{\linewidth}{@{}l l L@{}}
\toprule
Crop & Diagnosis & The follow-up it asks for \\
\midrule
Common bean & Bean anthracnose & close-up of affected leaves to check for accompanying foliar lesions and confirm anthracnose \\
Mango & White mango scale & close-up macro of the white encrustations to confirm individual scale insect bodies and check for crawlers \\
Maize & Fall armyworm & close-up of the whorl interior to confirm presence of larvae and characteristic inverted-Y head capsule markings \\
Citrus & Mealybug & macro close-up of a single infested stem to confirm individual mealybug bodies and waxy filaments versus fungal growth \\
Tomato & Leaf miner & close-up of a single affected leaf showing the serpentine mine trails to confirm leaf miner activity versus other foliar damage \\
\bottomrule
\end{tabularx}
\tnote{Quoted as written by the model. Each names a part of the plant, a sign to look for, and the alternative it would rule out.}
\end{table*}

\subsubsection{Advantages and Limitations}
\begin{itemize}
  \item \textbf{Advantages.} Single pass with no hand-off between stages; asks for a better photograph instead of guessing; extracts unstructured detail alongside the structured fields; one model to deploy.
  \item \textbf{Limitations.} Needs a GPU, which makes it the costlier route to host (Table~\ref{tab:tradeoff}), and at the slow end of the throughput we assume the two GPUs only just clear the busiest minute we measured, so more farmer queries need more GPUs (\S\ref{sec:res-routes}); disease accuracy is confounded by the stored production answer sitting in its training labels (\S\ref{sec:res-error}).
\end{itemize}

\subsection{Route B: Computer Vision Model Orchestration}
\label{sec:route-b}

\begin{figure*}[t]
\centering
\includegraphics[width=\linewidth]{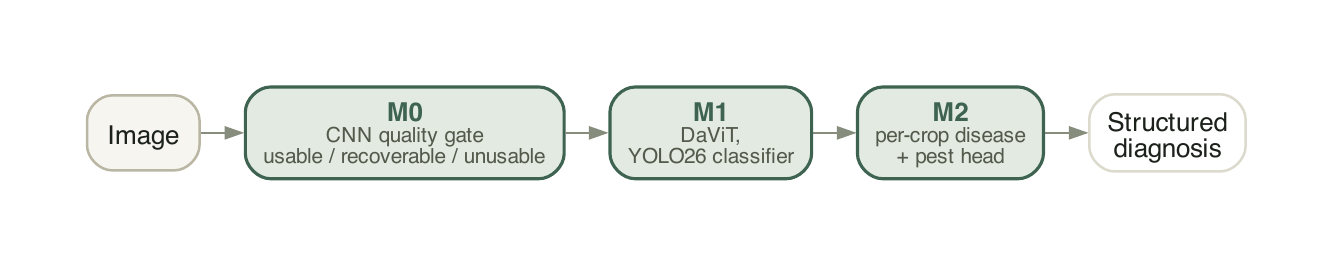}
\caption{Route B. Three independently versioned models.}
\label{fig:routeb}
\end{figure*}

Figure~\ref{fig:routeb} shows the route: one specialist model per stage, orchestrated. Each stage is versioned on its own, so a weak stage can be replaced without retraining the others. The price is a hand-off between each pair of stages. Table~\ref{tab:routeb} lists the candidate per stage.

\begin{table*}[t]
\centering\tabsize
\caption{Route B candidates by stage.}
\label{tab:routeb}
\begin{tabularx}{\linewidth}{@{}l L L@{}}
\toprule
Stage & Candidate & Role in the route \\
\midrule
M0, quality & MobileNetV3-small CNN & The learned gate of \S\ref{sec:m0-gateb}, chosen on its agreement with the paid gate \\
M1, crop (reference model) & DaViT-Base hierarchical crop head & The crop head of the four-head model, which also serves M2 below, so one model covers both stages \\
M1, crop (next-generation candidate) & YOLO26x-cls, fine-tuned & The same four-head design on a second backbone, scored against the reference model in \S\ref{sec:res-hier} \\
M2, disease & DaViT-Base hierarchical head, crop-masked & Picks the disease from the list allowed for the crop M1 predicted \\
M2, pest & Unfiltered pest head (\S\ref{sec:m2-routing}), unmasked 92-way, inside the same reference model & Names the insect from the full list, so a wrong M1 call cannot rule the right insect out \\
\bottomrule
\end{tabularx}
\tnote{Sections~\ref{sec:res-m0}, \ref{sec:res-crop} and \ref{sec:res-hier} score these candidates.}
\end{table*}

\begin{itemize}
  \item \textbf{Advantages.} Every threshold is explicit and independently tunable; a wrong answer traces to one stage; the models are small enough for CPU-only serving, with a mobile-viable quality gate; the lower hosting cost of the two routes (Table~\ref{tab:tradeoff}).
  \item \textbf{Limitations.} Three models to version, monitor and keep in sync; a wrong M1 crop call misroutes M2, so error compounds down the chain; no reasoning layer and no follow-up question; more moving parts to build and maintain than one model.
\end{itemize}

\section{Data and Methodology}
\label{sec:data}

\begin{figure*}[t]
\centering
\includegraphics[width=\linewidth]{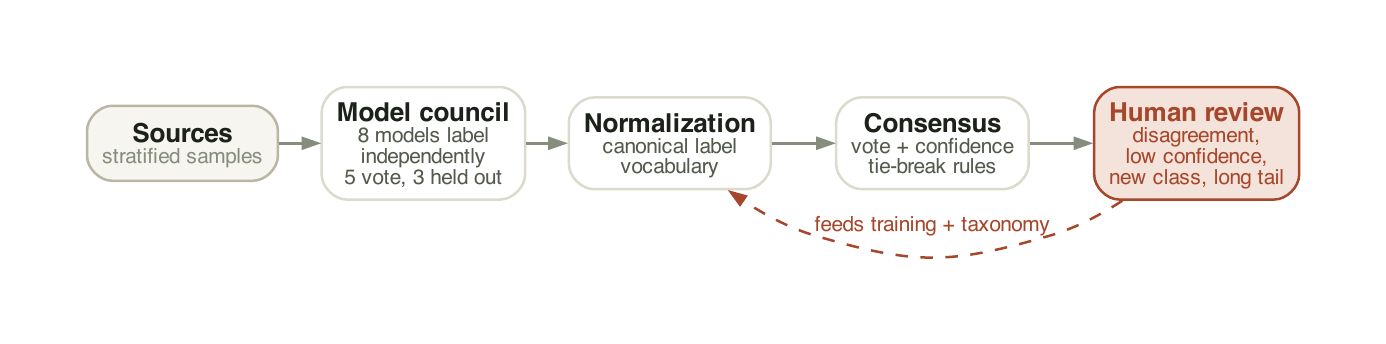}
\caption{Data curation path.}
\label{fig:data}
\end{figure*}

The labels are made by models, not by people labelling at scale. Plantix cannot supply them, because Plantix is one of the things being measured. Figure~\ref{fig:data} shows how the labels are built: a council of models labels every image on its own, normalisation and a consensus vote turn those into one label per image, and human review feeds back into both training and the label list.

\subsection{Sources and Sampling}
The images come from Farmer.Chat production data. Each image has a diagnosis label from Plantix, our production diagnosis service. These labels are used to stratify the initial sample by crop and by healthy, disease, and pest categories (Table~\ref{tab:splits}).

Plantix data or models are not used for training. The training labels are generated by the model council (\S\ref{sec:data-council}), with the production Plantix label used as the reference when the council does not reach agreement. Human review is the only non-model source of labels.

\subsection{Model Council}
\label{sec:data-council}
Eight models label every image on their own. Five of them vote on the final label. Three are kept out of the vote so that no model being tested helps write the label it is scored against. These are Plantix and the two models we fine-tune. Table~\ref{tab:panel} gives how often each model answered, and how often it agreed with the council, for crop and for disease.

\begin{table*}[t]
\centering\tabsize
\caption{The eight-model label council.}
\label{tab:panel}
\begin{tabular}{@{}llrrrr@{}}
\toprule
Model & Role & Crop answered & Crop agreement & Disease answered & Disease agreement \\
\midrule
Sonnet 4.6 & panel & 100.0\% & 82.0\% & 71.9\% & 68.3\% \\
GPT-5.4 & panel & 99.8\% & 81.2\% & 55.1\% & 65.7\% \\
Opus 4.8 & panel & 98.9\% & 80.9\% & 35.1\% & 72.2\% \\
Kimi K2.5 & panel & 90.9\% & 80.5\% & 59.8\% & 77.9\% \\
Pixtral Large & panel & 76.0\% & 68.8\% & 49.6\% & 77.2\% \\
Qwen3-VL, off the shelf & held out & 99.5\% & 78.4\% & 75.1\% & 69.8\% \\
Plantix & held out & 94.7\% & 80.5\% & 66.7\% & 58.8\% \\
Gemma-3 & held out & 65.0\% & 72.1\% & 54.6\% & 61.4\% \\
\bottomrule
\end{tabular}
\tnote{Crop columns are over the 114,721 rows with a council crop, disease columns over the 80,398 with a council disease. Whole labelled sample, not the test split. Agreement is a synonym match, the same map applied to the council label and to the answer as in \S\ref{sec:method}, counted over the rows the model answered: Opus 4.8 names one disease in three and agrees on three quarters of those. Plantix answers crop as a candidate list; one candidate counts as an answer, several do not.}
\end{table*}

\subsection{Normalisation and Consensus}
\label{sec:data-norm}
Eight models answer in free text, so the same plant can be described in eight different ways. Each answer is mapped onto the one label vocabulary of Table~\ref{tab:probtype} by a fixed set of rules, run as a script so the mapping can be repeated. Table~\ref{tab:norm} gives the rules with real answers from the data.

\begin{table*}[t]
\centering\tabsize
\caption{Normalisation rules, with answers taken from the model council's own output.}
\label{tab:norm}
\begin{tabularx}{\linewidth}{@{}L L L@{}}
\toprule
Rule & Answers as written & Becomes \\
\midrule
Plurals and spellings & tomatoes, tomato plant & tomato \\
Filler words & leaf miner damage, leaf miners, leaf miner feeding damage & leaf miner \\
Growth stage & maize seedling, coffee plant & maize, coffee \\
Scientific name & huanglongbing, helicoverpa armigera & citrus greening, cotton bollworm \\
Regional or trade name & panama disease, jassid, corn, paddy rice & fusarium wilt, leafhopper, maize, rice \\
Opposites kept apart & early blight, late blight; tree tomato, tomato & left as four separate classes \\
Model declines to answer & No crop detected, not defined, unidentified crop & \texttt{\_\_unspecified\_\_} \\
Family rather than crop & cucurbit, legume, grass-family crop & kept, marked low specificity \\
Several candidates at once & {[}bean, gram, soybean, pea{]} & \texttt{\_\_multi\_\_}, each member recorded \\
\bottomrule
\end{tabularx}
\tnote{Drawn from 922,976 answers (115,372 images by eight models). A refusal and a family-level answer are each kept as their own outcome, not discarded.}
\end{table*}

Two rules carry most of the work. Dropping filler (\emph{damage}, \emph{infestation}, \emph{feeding}, \emph{complex}) collapses the many ways a model describes the same insect, which is why \emph{thrips} absorbs six written forms. Against that, a guard list of words that change the meaning (\emph{early}, \emph{late}, \emph{downy}, \emph{powdery}, \emph{bacterial}, \emph{tree}) stops the same rule merging two real classes.

The final label is then a vote, with confidence and tie-break rules. Disagreement, low confidence, new classes and rare classes go to a human reviewer. The vote often produces no diagnosis: more than half of all disease answers are refusals to name the problem, so many images have no council label to count.

\subsection{Dataset Splits}
\label{sec:method-splits}
Table~\ref{tab:splits} gives the two modules trained here. Route~A trains on its own draw from the same pool, and \S\ref{sec:results} names the row set behind every results table.

\begin{table}[tb]
\centering\tabsize
\caption{Images used to train, validate and test each module.}
\label{tab:splits}
\begin{tabular}{@{}l r r r@{}}
\toprule
Module & Train & Validation & Test \\
\midrule
M0, quality gate & 19,994 & 3,000 & 4,999 \\
M1, crop & 82,991 & 9,220 & 16,275 \\
\bottomrule
\end{tabular}
\tnote{The M0 splits never share an image identifier. The M1 split is the hierarchical manifest, so the same images train the M2 disease and pest heads.}
\end{table}

Two experiments run side by side. Table~\ref{tab:single} scores the single scoring dataset. The hierarchical test (Table~\ref{tab:hier}) is a separate experiment with its own four-part label set, built to answer the disease-against-pest question that one diagnosis grade cannot. No number from one is compared with a number from the other.

\subsection{Scoring Rule}
\label{sec:method}
One rule scores every system, in three steps:
\begin{enumerate}
  \item \textbf{Clean both labels.} Lower-case them, cut any free text after a dash, colon, comma or bracket, and apply the same synonym map to the reference and the prediction. The map merges names for one crop or problem that the label list holds twice, such as sugar beet and beet, so a system is not marked wrong for choosing the other spelling.
  \item \textbf{Count a non-answer as a wrong answer.} Markers such as unspecified, and answers listing several candidates, become ``no answer'' and score as wrong. The row set is the same for every system, so no system can raise its score by answering less often.
  \item \textbf{Score twice.} The strict column counts only an exact match. The containment column is more generous: it counts an answer when either label contains the other (``mosaic virus'' against ``cucumber mosaic virus''), which accommodates models that answer in their own words.
\end{enumerate}
Two kinds of row stay outside the diagnosis count for every system alike: references that are healthy or unspecified, and references that are Plantix's own stored answer. Dropping the second kind stops Plantix grading itself. It does not do the same for GPT-5.4, which voted on the panel consensus (Table~\ref{tab:panel}); \S\ref{sec:res-crop} names that exposure where its column is read.

\subsection{Baselines and Systems Compared}
Table~\ref{tab:baselines} defines the systems compared throughout the paper.
\begin{table*}[t]
\centering\tabsize
\caption{Baseline and proposed system definitions used throughout the paper.}
\label{tab:baselines}
\begin{tabularx}{\linewidth}{@{}l L@{}}
\toprule
ID & Definition \\
\midrule
Baseline 0 & Existing production pipeline (GPT-4o quality gate and Plantix diagnosis call) \\
Baseline 1 & General-purpose VLM, no fine-tuning (Gemini 3.5 Flash reported; GPT-5.4, Claude and Kimi scored as labelling-panel members) \\
Route A & Fine-tuned vision-language model, single call, all three stages \\
Route B & Specialist computer-vision models, one per stage, orchestrated \\
\bottomrule
\end{tabularx}
\end{table*}

\subsection{Pipeline Configurations}
Table~\ref{tab:configs} lists the configurations under test.
\begin{table*}[t]
\centering\tabsize
\caption{Configurations under test.}
\label{tab:configs}
\begin{tabularx}{\linewidth}{@{}>{\raggedright\arraybackslash}p{3.6cm} >{\raggedright\arraybackslash}p{2.2cm} >{\raggedright\arraybackslash}p{3.2cm} L@{}}
\toprule
Configuration & M0 & M1 & M2 \\
\midrule
Baseline 0, existing & Existing & Existing & Existing \\
Baseline 1, general VLM & VLM & VLM & VLM \\
Route A, fine-tuned VLM & VLM & VLM & VLM \\
Route B, CV orchestration & CNN gate & DaViT, YOLO26 cls & Hierarchical disease and pest heads \\
\bottomrule
\end{tabularx}
\end{table*}
Two comparisons follow from it: the existing gate against the proposed M0 (Table~\ref{tab:gateb}), and one model doing everything against the three-stage pipeline (\S\ref{sec:res-crop}, with Route~B's per-stage numbers in \S\ref{sec:res-hier}). Two more are listed in \S\ref{sec:limits}: M0 with and without the cleaning step, and training with and without human-reviewed labels.

\subsection{Metrics by Stage}
Table~\ref{tab:metrics} lists the metrics reported per stage.
\begin{table*}[t]
\centering\tabsize
\caption{Metrics reported, one row per metric.}
\label{tab:metrics}
\begin{tabularx}{\linewidth}{@{}l L l@{}}
\toprule
Metric & What it measures & Stage \\
\midrule
Agreement with the reference gate & How often the small gate makes the same call as the paid gate, overall and per check & M0 \\
p50 latency & Median time to judge one image & M0 \\
Model size & File size, which decides whether the model fits on a phone & M0 \\
Crop accuracy & Share of images whose crop is named correctly & M1 \\
Declined share & Share of images where the model names no crop & M1 \\
Diagnosis accuracy & Share of images whose problem is named correctly, exactly and with containment credit & M2 \\
Head accuracy & The same, one score per head: crop, category, disease, pest & M2 \\
Disease macro-F1 & Every disease class scored equally, so rare ones count as much as common ones & M2 \\
Monthly hosting cost & Cost of running the route at the demand we measured & System \\
Deployability & What hardware the route needs & System \\
\bottomrule
\end{tabularx}
\end{table*}
Section~\ref{sec:limits} lists two further measurements that follow this paper: useful-image recall and false rejections at M0, and confidence calibration at M1.

\section{Results}
\label{sec:results}

\subsection{Component-Level Results: M0}
\label{sec:res-m0}
\begin{figure*}[t]
\centering
\includegraphics[width=\linewidth]{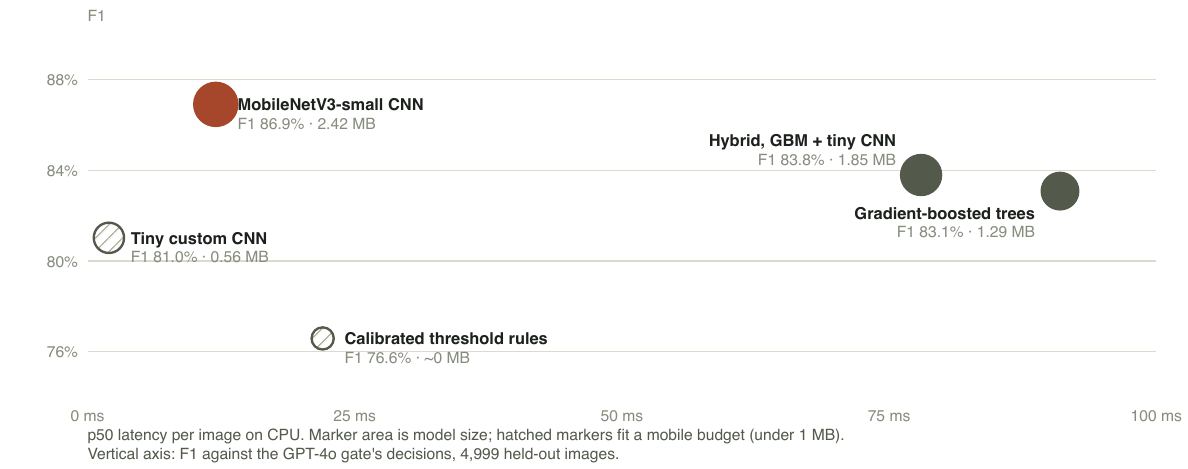}
\caption{Gate B candidates, F1 against p50 latency on the held-out test.}
\label{fig:m0}
\end{figure*}

Table~\ref{tab:gateb} and Figure~\ref{fig:m0} score the five Gate~B candidates of \S\ref{sec:m0-gateb} against Gate~A's answers on the 4,999-image held-out test of Table~\ref{tab:splits}, on agreement, size and latency. Table~\ref{tab:percheck} gives the per-check detail for the winner.

\begin{table*}[t]
\centering\tabsize
\caption{Gate B candidates on the 4,999-image held-out test, scored against Gate A.}
\label{tab:gateb}
\begin{tabular}{@{}lrrrrl@{}}
\toprule
Candidate & F1 & Accuracy & Size & p50 latency & Fits mobile ($<$1 MB) \\
\midrule
\textbf{MobileNetV3-small CNN} & \best{86.91\%} & \best{86.76\%} & 2.42 MB & 12 ms & \no \\
Hybrid (GBM + tiny CNN) & 83.79\% & 83.88\% & 1.85 MB & 78 ms & \no \\
Gradient-boosted trees & 83.08\% & 82.90\% & 1.29 MB & 91 ms & \no \\
Tiny custom CNN & 81.02\% & 82.52\% & 0.56 MB & \best{2 ms} & \yes \\
Calibrated threshold rules & 76.58\% & 77.32\% & $\sim$0 MB & 22 ms & \yes \\
\bottomrule
\end{tabular}
\tnote{Green bold marks the best F1, accuracy and latency. A tick in the last column means the model fits inside a 1 MB on-device budget.}
\end{table*}

\begin{table}[tb]
\centering\tabsize
\caption{Per-check F1 for MobileNetV3-small, ranked.}
\label{tab:percheck}
\begin{tabular}{@{}lr@{}}
\toprule
Check & F1 \\
\midrule
Plant Detected & \best{92.2\%} \\
Dominant Content & 88.2\% \\
Orientation & 80.8\% \\
Obstruction & 80.2\% \\
Noise & 79.6\% \\
Color Balance & 79.0\% \\
Lighting & 77.8\% \\
Motion Blur & 76.8\% \\
Resolution & 76.2\% \\
Focus & 74.8\% \\
\bottomrule
\end{tabular}
\tnote{Green bold marks the best per-check F1.}
\end{table}

\begin{itemize}
  \item MobileNetV3-small wins at 86.91\% F1 and 12~ms; the tiny CNN reaches 81.02\% at 2~ms and is the only learned candidate that fits on a phone. Recommendation: MobileNetV3 for server deployment, the tiny CNN for on-device use.
  \item Threshold rules alone trail every learned candidate (Table~\ref{tab:gateb}). A learned model is required, not only calibration.
  \item No candidate reaches the 88\% target of \S\ref{sec:m0-gateb}; the best is 1.1 points short. That may be close to the limit imposed by Gate~A's own inconsistency.
  \item The four checks about what is in the picture are also the four on which the small model has the highest F1 (Table~\ref{tab:percheck}), led by Plant Detected and Dominant Content. Dominant Content is also the check that fails most often (Table~\ref{tab:qfail}).
\end{itemize}

\subsection{Crop and Diagnosis on One Test Set}
\label{sec:res-crop}
Table~\ref{tab:single} is the headline comparison. Seven systems are scored on the same 10,335 images under the one rule of \S\ref{sec:method}: the Route~A fine-tune, the same model before fine-tuning, the two Route~B backbones, the production baseline, a panel model (GPT-5.4) and a general-purpose baseline. Table~\ref{tab:splits} gives the row counts.

Two systems are left out: the Gemma-3 fine-tune and the model it started from, both rejected candidates (\S\ref{sec:route-a}).

\begin{table*}[t]
\centering\tabsize
\caption{Seven systems on one test set, scored under one rule.}
\label{tab:single}
\begin{tabularx}{\linewidth}{@{}L L r r r r@{}}
\toprule
System & Role & Crop & Diagnosis & Containment & No answer \\
\midrule
\textbf{DaViT-Base, hierarchical} & Route B reference model & \best{95.41\%} & \best{73.26\%} & \best{77.43\%} & \best{0\%} \\
Qwen3-VL-4B, fine-tuned & Route A, out of sample & 94.66\% & 33.04\% & 36.97\% & 41.30\% \\
YOLO26x-cls, hierarchical & Route B next-generation candidate & 93.78\% & 68.42\% & 72.63\% & 0\% \\
Plantix & production baseline & 91.46\% & 42.61\% & 53.50\% & 35.99\% \\
GPT-5.4 & panel member & 88.25\% & 40.36\% & 45.86\% & 42.47\% \\
Gemini 3.5 Flash & general-purpose baseline & 86.16\% & 39.05\% & 46.95\% & 14.67\% \\
Qwen3-VL-4B, off the shelf & Route A before fine-tuning & 73.34\% & 12.96\% & 16.03\% & 45.08\% \\
\bottomrule
\end{tabularx}
\tnote{Crop is over all 10,335 rows; diagnosis, containment and no-answer over the 7,636 whose reference came from the panel. Green bold marks the best per column.}
\end{table*}

Scoring rules: a refusal counts as a miss, containment is the generous match of \S\ref{sec:method}, and backbone rows are the mean of two seeds. GPT-5.4 voted on the panel labels (Table~\ref{tab:panel}), so it helped write the reference that its own diagnosis column is scored against.

Four readings of Table~\ref{tab:single}:
\begin{itemize}
  \item \textbf{The set is every top-20 crop test row on which every scored system has a prediction on record.} The production crop classifier determined the usable set, declining 12.5\% of the 11,810 top-20 test rows and leaving the 10,335 scored here.
  \item \textbf{The two families fail differently.} Neither backbone ever refuses to answer. The language models return no diagnosis on 15\% to 45\% of panel-labelled rows, and Plantix on 36\%. Being generous about wording does not close the gap: DaViT-Base leads Plantix by 30.7 points on exact matches and by 23.9 when either label containing the other counts.
  \item \textbf{Route B names the crop best.} The fine-tune is trained on rows disjoint from this test set and reaches 94.66\%, below DaViT-Base. It learned from the same panel that wrote the reference labels, so part of its score may reflect agreement with that panel's naming conventions.
  \item \textbf{The backbones answer from a fixed list} of 110 crops and 285 diagnoses; the language models answer in their own words, which is what the containment column allows for (\S\ref{sec:method}).
\end{itemize}

\subsection{Route A Against Route B}
\label{sec:res-routes}
\begin{table*}[t]
\centering\tabsize
\caption{Route A against Route B.}
\label{tab:tradeoff}
\begin{tabularx}{\linewidth}{@{}l L L@{}}
\toprule
Dimension & Route A, fine-tuned VLM & Route B, CV orchestration \\
\midrule
Crop accuracy & 94.66\%, out of sample & \best{95.41\%} reference model, 93.78\% next-generation candidate \\
Diagnosis on panel-labelled rows & 33.04\%, label-confounded & \best{73.26\%} reference model, 68.42\% next-generation candidate \\
Monthly hosting cost, measured demand & \$924 (two g6.xlarge, L4 GPU, one-year reservation) & \best{\$100} (two c7g.xlarge, CPU only, one-year reservation) \\
Follow-up question & Asks for a second photograph on 57.5\% of rows (\S\ref{sec:res-error}) & not available \\
Deployability & GPU required & CPU only; on-device quality gate possible \\
Failure mode & one wrong answer, no chain & a wrong M1 call misroutes M2 \\
\bottomrule
\end{tabularx}
\tnote{Crop and diagnosis from Table~\ref{tab:single}, cost from the hosting model below. Green bold marks the better route where the two are comparable.}
\end{table*}

These costs are modelled, not billed:
\begin{itemize}
  \item Prices come from the AWS list published 10 August 2026 for ap-south-1, applied to the query volume we measured: 46,869 diagnosis calls in the month to 24 July 2026.
  \item The CPU route's speed comes from measured laptop figures. The GPU route's speed, 0.5 to 5 images per second per GPU, is assumed with no measurement behind it. This is the least certain assumption in the \$924 estimate.
  \item Most of either bill is idle time: under 8\% of the paid hours are used in every scenario, and even at the slow end of its measured range the CPU route provides 14.6$\times$ the capacity required for the busiest minute we measured, while the GPU route at its slowest assumed speed only just meets that demand.
\end{itemize}

\subsection{Country and Crop Analysis}
\label{sec:res-country}
\begin{itemize}
  \item India has 45.0\% of its submissions rejected before diagnosis, compared with 18.3\% in Kenya (Table~\ref{tab:funnel}). A single global quality threshold under-serves one country or the other.
  \item Wheat and maize are 41.8\% of crop-named images, and the top 20 crops cover 89.2\% (Figure~\ref{fig:coverage}).
  \item Of the 20 reference crops, 17 reach 80\% or better for the Route~A fine-tune. Cotton (72.8\% of 320 rows) and cucumber (76.0\% of 495) are the exceptions; onion has only two test rows, so its accuracy is not informative.
\end{itemize}

\subsection{Error Analysis}
\label{sec:res-error}
Table~\ref{tab:confusion} lists the crop pairs that the Route~A fine-tune misclassifies most often, on the same 10,335 rows.
\begin{table}[tb]
\centering\tabsize
\caption{Crop pairs the Route~A fine-tune misclassifies most often.}
\label{tab:confusion}
\begin{tabular}{@{}llr@{}}
\toprule
Reference & Predicted & Share of crop errors \\
\midrule
Cucumber & Cucurbit & 13.9\% \\
Wheat & Grass-family crop & 6.9\% \\
Soybean & Common bean & 3.6\% \\
Cotton & Okra & 3.4\% \\
Maize & Grass-family crop & 3.4\% \\
\bottomrule
\end{tabular}
\tnote{Shares are of the model's 552 crop errors on the 10,335 scored rows, which are 5.3\% of those rows.}
\end{table}

\begin{itemize}
  \item Nearly a third of its crop errors, 29.5\%, name a family rather than a crop: cucurbit for cucumber, grass-family crop for wheat or maize. Those errors reflect an overlap in the label list, not a misreading of the image.
  \item The fine-tune asks for a second photograph on 57.5\% of the 10,335 rows, and on 1,288 of the 2,452 panel-labelled rows where it had already named both the crop and the problem correctly.
\end{itemize}

\subsection{Backbone Benchmark: DaViT-Base Against YOLO26x-cls}
\label{sec:res-hier}
A held-out test of 16,273 images. The disease filter uses the model's own crop prediction, which is what a deployed system would have, not the true crop. Both backbones were trained the same way, 25 epochs on the hierarchical manifest of Table~\ref{tab:splits}. Values are the mean of two seeds (42 and 1337), and the two seeds differ by less than a point on every head.

\begin{table*}[t]
\centering\tabsize
\caption{Four-head accuracy on the held-out test, mean of two seeds.}
\label{tab:hier}
\setlength{\tabcolsep}{4pt}\begin{tabular}{@{}lrrrrrrr@{}}
\toprule
Backbone & Params & Mean head acc.\ & Crop & Category & Disease & Pest & Disease macro-F1 \\
\midrule
\textbf{DaViT-Base} & 87.4M & \best{82.65\%} & \best{89.83\%} & \best{89.27\%} & \best{77.86\%} & \best{73.66\%} & \best{41.75\%} \\
YOLO26x-cls & 29.0M & 79.78\% & 87.75\% & 87.02\% & 74.14\% & 70.21\% & 39.03\% \\
\bottomrule
\end{tabular}
\tnote{Crop (110 choices) and category (3) are over all 16,273 rows; disease (285, filtered by predicted crop) over 11,916; pest (92, unfiltered) over 4,276.}
\end{table*}

\begin{table*}[t]
\centering\tabsize
\caption{Training and inference cost of the same comparison.}
\label{tab:hiercost}
\begin{tabular}{@{}lrrrr@{}}
\toprule
Backbone & Train img/s & Train wall (2 seeds) & Peak GPU memory & Eval img/s \\
\midrule
DaViT-Base & 1,430 & 54.4 min & 18.97 GB & 655 \\
YOLO26x-cls & \best{3,699} & \best{23.3 min} & \best{17.31 GB} & \best{1,197} \\
\bottomrule
\end{tabular}
\tnote{Four L40S GPUs, distributed data parallel. Train wall is the measured elapsed time summed over the two seed runs. Green bold marks the better value per column.}
\end{table*}

\begin{figure*}[t]
\centering
\includegraphics[width=\linewidth]{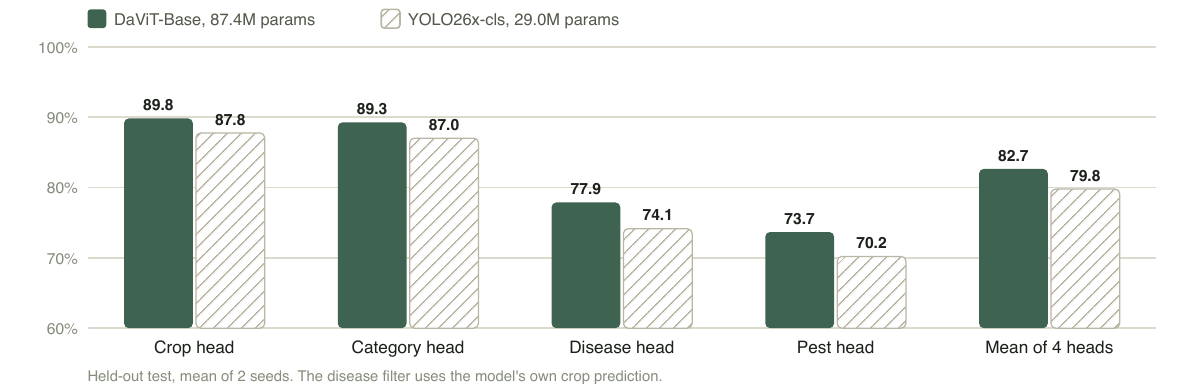}
\caption{Head accuracy for the two backbones, mean of two seeds.}
\label{fig:heads}
\end{figure*}

\begin{itemize}
  \item Mean head accuracy is the plain average of the four heads, which do not share a denominator.
  \item DaViT-Base wins every head by 2.1 to 3.7 points (Figure~\ref{fig:heads}), but the two differ by 3.0$\times$ in parameter count, so part of the gap is capacity rather than design.
  \item YOLO26x-cls trains 2.3$\times$ faster (11.7 against 27.2 minutes per run, Table~\ref{tab:hiercost}) and has about 1.8$\times$ higher inference throughput, with one third as many parameters.
  \item The disease result is not a 285-class classification. The crop filter cuts it to a median of 10 allowed diseases per crop (lowest 8, highest 43).
  \item The pest head is never filtered by crop, by design. It scores below disease (70.21\% to 73.66\% against 74.14\% to 77.86\%), which is the expected price of picking from all 92 pests with no help from the crop.
  \item The labels come from the council's vote, with the stored production answer standing in where the council did not agree, not from experts. So this table measures agreement with that consensus, the same caveat as \S\ref{sec:res-crop}.
\end{itemize}

\section{Impact and Future Work}
\label{sec:limits}

\subsection{Impact}
\label{sec:impact}
No route runs on live farmer queries at the time of writing, so nothing below is an observed
outcome. Table~\ref{tab:impact} compares the four systems that answer the crop and the diagnosis
question (M1 and M2) on the three deployment dimensions of cost, accuracy and speed.

\begin{table*}[t]
\centering\tabsize
\caption{Crop and diagnosis systems against cost, accuracy and speed.}
\label{tab:impact}
\begin{tabularx}{\linewidth}{@{}L l r r L@{}}
\toprule
System & Monthly cost & Crop & Diagnosis & Measured speed \\
\midrule
\textbf{DaViT-Base, hierarchical} & \best{\$100}, CPU & \best{95.41\%} & \best{73.26\%} & 655 img/s, batched \\
YOLO26x-cls, hierarchical & \best{\$100}, CPU & 93.78\% & 68.42\% & \best{1,197 img/s}, batched \\
Qwen3-VL-4B, fine-tuned & \$924, GPU & 94.66\% & 33.04\% & not measured \\
Plantix & paid per call & 91.46\% & 42.61\% & 2.8 to 4.4~s per image \\
\bottomrule
\end{tabularx}
\tnote{Accuracy from Table~\ref{tab:single}, cost from Table~\ref{tab:tradeoff}. Backbone speed is batched evaluation on four L40S GPUs, not the CPU shape the cost assumes, and the Plantix figure is a single-request production trace (Table~\ref{tab:existingfacts}), so the last column compares three different measurements. Green bold marks the best per column.}
\end{table*}

Three readings of Table~\ref{tab:impact}:
\begin{itemize}
  \item \textbf{Cost and accuracy do not present a trade-off here.} The cheaper route is also the more accurate one on both questions, so the case for the GPU route rests on what it does besides classify: the follow-up question and the free-form reasoning (\S\ref{sec:route-a}).
  \item \textbf{Speed is the weakest column and should not decide anything yet.} The two backbones are timed batched on training hardware, the production path is timed per request, and the fine-tune has no number at all. A single benchmark using the deployment configuration for each system would resolve this.
  \item \textbf{The gap the table does not show is control.} Every value in the Plantix row is fixed: no threshold to set, no crop to add, no confidence to read (Table~\ref{tab:existingfacts}). The two backbone rows provide control over all three.
\end{itemize}

Beyond crops, the three questions the pipeline asks, whether an image can be read, what is in it, and
what is wrong with it, carry to the condition of a farm animal, the grading of produce, and any image
task where whether the photograph is usable is a separate question from what it shows. The
photographs in this corpus that appear to show animals (Table~\ref{tab:funnel}, last column) are a
natural next application.

\subsection{Future Work}
\begin{enumerate}
  \item \textbf{False-rejection rate at M0.} How many usable images the gate rejects needs its own measurement, using a recall-versus-threshold sweep of Gate~B.
  \item \textbf{Image enhancement.} No cleaning method is evaluated.
  \item \textbf{Not modelled.} Disease severity, multi-disease images, multi-crop images, disease progression over time.
  \item \textbf{AI-aided human review.} Only a small set of independently reviewed images reaches training; reviewers will instead correct the pipeline's own answer under written guidelines (\S\ref{sec:fail-review}).
  \item \textbf{Confidence calibration.} M1 confidence will be calibrated, so a cut-off can be set from a target error rate.
\end{enumerate}

\section{Conclusion}
\label{sec:conclusion}
\begin{itemize}
  \item Diagnosing a crop from a field photograph is five separate decisions (\S\ref{sec:intro}), and the production system lets us adjust none of them.
  \item Measuring M0, M1 and M2 separately means every threshold is ours to set and every failure can be traced to one stage, which a single closed call cannot do.
  \item A small local MobileNetV3 matches the GPT-4o quality gate at 86.9\% F1 and 12~ms, 1.1 points under the 88\% target we set beforehand.
  \item On one test set of 10,335 images, scored the same way for every system, the DaViT-Base backbone achieves the highest crop accuracy at 95.41\%, 0.75 points above the fine-tuned VLM and 3.95 above Plantix, on images it never saw. It leads every system on diagnosis too: 73.26\%, against 68.42\% for the smaller YOLO backbone, 42.61\% for Plantix and 33.04\% for the fine-tune, at one ninth the VLM route's modelled hosting cost. No diagnosis number is final until the stored production answer is out of every prompt and training set, and until disease and pest are split into separate label lists.
  \item Route~A and Route~B sit at different points on accuracy, cost and control (Table~\ref{tab:tradeoff}). Which one to deploy is a product decision, and this paper supplies the measurements it needs.
\end{itemize}

\begingroup
\footnotesize
\setlength{\bibsep}{2pt plus 1pt}
\bibliographystyle{unsrtnat}
\bibliography{references}
\endgroup

\end{document}